\pdfoutput=1

\documentclass[11pt]{article}

\usepackage[]{emnlp2021}

\usepackage{times}
\usepackage{latexsym}

\usepackage[T1]{fontenc}

\usepackage[utf8]{inputenc}

\usepackage[letterspace=-26]{microtype}

\usepackage[utf8]{inputenc}
\usepackage{amsmath}
\usepackage{amssymb}
\usepackage{mathtools}
\usepackage{arydshln}
\usepackage{booktabs}
\usepackage{tabularx}
\usepackage{multirow}
\usepackage{tikz}
\usepackage{pgfplots}
\usepackage[normalem]{ulem}
\usepackage{xcolor}
\usepackage{dashbox}%
\usepackage{textcomp}
\usepackage{tcolorbox}
\usepackage{adjustbox}
\usepackage{lipsum}
\usepackage[inline]{enumitem}
\usepackage{relsize}

\usepackage{pmboxdraw}

\pgfplotsset{compat=1.13}

\definecolor{c0}{RGB}{75,122,149} 
\definecolor{c1}{RGB}{255,103,103} 
\definecolor{darkgrey}{RGB}{149,149,149}
\definecolor{decentgrey}{RGB}{242,242,242}

\tikzset{
	keep name/.style={
		prefix after command={
			\pgfextra{\let\fixname\tikzlastnode}
		}
	},
	partialbox/.style={
		keep name,
		append after command={
			(\fixname.north) -- 
			(\fixname.north west) -- 
			(\fixname.south west) -- 
			([xshift=-#1]\fixname.south)
			(\fixname.north) -- 
			(\fixname.north east) -- 
			(\fixname.south east) -- 
			([xshift=#1]\fixname.south)
		}
	},
	partialbox/.default=5pt
}

\usetikzlibrary{calc,fit,positioning,arrows,arrows.meta}

\pgfdeclarelayer{bg}
\pgfsetlayers{bg,main}

\newtcbox{\pattern}{on line,colback=c0!10,colframe=white,size=fbox,arc=3pt, box align=base,before upper=\strut,
top=-2pt, bottom=-2pt, boxrule=0pt}
\newtcbox{\inlinepattern}{on line,colback=c0!10,colframe=white,size=fbox,arc=3pt, box align=base,before upper=\strut,
	top=-4pt, bottom=-4pt, boxrule=0pt}
\newtcolorbox{multipattern}{on line,colback=decentgrey!75,colframe=white,size=fbox,arc=3pt, box align=base, top=0pt, bottom=2pt, boxrule=0pt, before=\adjustbox{valign=c}\bgroup, after=\egroup, before upper=\strut}

\newcolumntype{Y}{>{\centering\arraybackslash}X}

\newcommand{\pegasus}{\textsc{Pegasus}}
\newcommand{\pet}{\textsc{Pet}}
\newcommand{\genpet}{\textsc{genPet}}

\newcommand\mask{\_\_}

\newcommand{\bt}{\fontseries{b}\selectfont}

\newcommand\p{\phantom{0}}

\title{Few-Shot Text Generation with Natural Language Instructions}

\author{
	Timo Schick \and Hinrich Sch\"utze \\[0.5em]
	Center for Information and Language Processing, LMU Munich, Germany \\[0.5em]
	{\tt schickt@cis.lmu.de}
}

\date{}

\newcounter{notecounter}

\newcommand{\enoteson}{\long\gdef\enote##1##2{{
\stepcounter{notecounter}
{\large\bf
\hspace{1cm}\arabic{notecounter} $<<<$ ##1: ##2
$>>>$\hspace{1cm}}}}}
\enoteson

\begin{document}
\maketitle

\begin{abstract}
	
	Providing pretrained language models with simple task descriptions in natural language enables them to solve some tasks in a fully unsupervised fashion.
	Moreover, when combined with regular learning
	from examples, this idea yields impressive few-shot
	results for a wide range of text classification
	tasks. It~is also a promising direction to improve data efficiency in
	generative settings, but there are several
	challenges to using a combination of task
	descriptions and example-based learning for text
	generation. In particular, it is crucial to find
	task descriptions that are easy to understand for the pretrained model and to ensure that it actually makes good use of them; furthermore, effective measures against overfitting have to be implemented.
	In this paper, we show how these challenges can be
	tackled: We introduce \genpet{}, a method for text generation that is based on \emph{pattern-exploiting training}, a recent approach for combining textual instructions with supervised learning that only works for classification tasks.
	On several summarization and headline generation datasets, \genpet{} gives consistent improvements over strong baselines in few-shot settings.\footnote{Our implementation of \genpet{} and code to recreate our few-shot training datasets is publicly available at \url{https://github.com/timoschick/pet}.}
	
\end{abstract}

\section{Introduction}
\label{intro}
Pretraining large neural networks with a language modeling objective has led to significant improvements throughout NLP \citep[][\emph{i.a.}]{peters2018deep,howard2018universal,radford2018improving,devlin2018bert,raffel2019exploring,brown2020language}. Further improvements are often possible by choosing a different pretraining objective that more closely matches the downstream task of interest. Examples include casing prediction for named entity recognition \citep{Mayhew_Nitish_Roth_2020}, gap sentence generation for summarization \citep{zhang2019pegasus}, and sentence unshuffling for discourse representations \citep{lee2020slm}.

\begin{figure}
	\tikzset{
		every node/.style={
			outer sep=0, font=\sffamily\footnotesize
		},
		pattern/.style={
			text height=1.5ex, fill=c0!10, rounded corners=3pt, outer sep=2pt, inner ysep=4pt, text width=3cm, inner xsep=2pt, align=center,
		},
		output/.style={
			text height=1.5ex, text width=3.7cm, outer sep=2pt, rounded corners=3pt, fill=c0!10, inner ysep=6pt, inner xsep=6pt,
		},
		arrow/.style={
			->,>={Latex[length=1mm, width=1mm]}, draw=darkgrey
		},
	}
	\centering
	\begin{tikzpicture}
	\node[output](p0-out){Please contact us if you have any questions.};	
	\node[pattern, left=0.2cm of p0-out](p0){{\color{c0}\textbf{x}} \mask{}};
	\path[] (p0) edge[arrow] (p0 -| p0-out.west);	
	
	
	\node[output, below=0.25cm of p0-out](p2-out){Your Internet Banking accounts are now setup again for accessing.};	
	\node[pattern, left=0.2cm of p2-out](p2){{S\hspace*{-0.5pt}h\hspace*{-0.5pt}o\hspace*{-0.5pt}r\hspace*{-0.5pt}t\hspace*{-0.5pt} S\hspace*{-0.5pt}u\hspace*{-0.5pt}m\hspace*{-0.5pt}m\hspace*{-0.5pt}a\hspace*{-0.5pt}r\hspace*{-0.5pt}y: \hspace*{-0.8pt}\mask{}} \color{c0}\textbf{x}};
	\path[] (p2) edge[arrow] (p2 -| p2-out.west);
	
	\node[output, below=0.25cm of p2-out](p3-out){Internet Banking Password reset?};	
	\node[pattern, left=0.2cm of p3-out](p3){{E-Mail Title: \mask{}} \color{c0}\textbf{x}};
	\path[] (p3) edge[arrow] (p3 -| p3-out.west);
	
	\begin{pgfonlayer}{bg}
	\node[inner sep=0, outer sep=0] at (p0-out.north -| p2.center) (pos-helper-1){};		
	\node[inner sep=0, outer sep=0] at (p0-out.north -| p0-out.center) (pos-helper-2){};
	\node[above=0cm of pos-helper-1](){\textbf{Instructions}};
	\node[above=0cm of pos-helper-2](){\textbf{Generated Texts}};
	\end{pgfonlayer}
	
	\end{tikzpicture}
	\caption{Texts generated by \pegasus{}-large with different instructions for input {\footnotesize\sffamily\color{c0} \textbf{x} = Dear John, Your Internet Banking accounts are now setup again for accessing. The login id is still your main account with the password being reset to the last six (6) digits of your SSN.} Without any instructions, the model simply generates a continuation of the given input (top). Providing an instruction makes it generate an appropriate summary (center) or e-mail title (bottom) even in zero-shot settings and enables much more data-efficient learning.}
	\label{figure:motivational-example}
\end{figure}
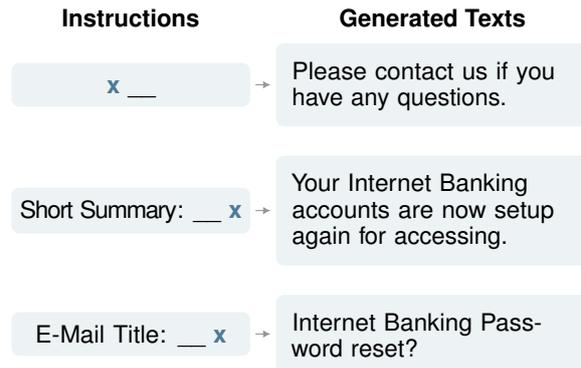

While such approaches can significantly reduce the amount of
training data required, they typically still do not perform
well if only a handful of examples is available for the
downstream task, which is a common scenario for many
real-word uses of NLP. In such few-shot settings, however,
significant gains are possible by
reversing what is adapted to what:
Instead of making
pretraining more similar to a downstream task, we can
reformulate the downstream task to make it more similar to the
pretraining objective. For masked language
models \citep[e.g.,][]{devlin2018bert,lewis2019bart}, one
such reformulation technique
is to convert inputs to cloze
questions by adding a text snippet that contains some form
of task description, often in the form of a short
prompt \citep{radford2018language,schick2020exploiting}. 
Besides making pretraining and finetuning more similar, this approach has the compelling benefit of enabling users to \emph{explain} a task to a pretrained model, making it much easier for the model to understand the task.
This is illustrated
in Figure~\ref{figure:motivational-example}, where a pretrained language model is given the same input with different instructions and adapts its output accordingly.

The idea of providing task descriptions even works in an unsupervised
setting \citep{radford2018language} or when examples are
simply provided as additional
context \citep{brown2020language}; however, it only unfolds its full
potential when combined with gradient-based training on a
handful of labeled examples \citep{schick2020just}. Unfortunately, current approaches for doing so 
are limited to text \emph{classification} tasks \citep{schick2020exploiting}.
Inspired by their success, we investigate whether the underlying idea 
can also be transferred to more challenging text-to-text tasks that require the
\emph{generation} of text sequences given an input text, such as abstractive summarization.
We introduce
\genpet{},
a novel method based on \pet{} \citep{schick2020exploiting}, that enables 
finetuning of generative language models using
both instructions and labeled examples.
We show that \genpet{} is a highly data-efficient method that enables us
to finetune a pretrained \pegasus{}
model \citep{zhang2019pegasus} with as little as 10 or 100 training examples. We evaluate our
approach on a diverse set of six English headline generation
and text summarization tasks both in zero-shot and few-shot
settings and show that \pegasus{} trained with \genpet{} clearly outperforms regular finetuning.

In summary, our contributions are as follows:

\begin{itemize}[topsep=0.5em]
	\setlength\itemsep{-0.1em}
	\item We introduce \genpet{}, a finetuning procedure for generative language models that achieves great data efficiency by using both textual instructions and training examples.
	\item We show that training \pegasus{} with \genpet{} outperforms standard finetuning across a broad set of tasks and training set sizes.
	\item We analyze the factors contributing to \genpet{}'s strong performance and quantify the impact of all its components.
\end{itemize}

\section{Related Work}
\label{sec:related-work}

Masked language modeling was proposed as a pretraining
objective by \citet{devlin2018bert}. Several variants of this objective
that involve generating sequences of
text have been proposed, including T5 \citep{raffel2019exploring}, \textsc{Bart} \citep{lewis2019bart} and \pegasus{} \citep{zhang2019pegasus}, of which we make use in this work.

The idea to rephrase tasks as cloze questions is commonly used to probe the knowledge contained within masked language models \citep[e.g.,][]{Petroni_2019,wang-etal-2019-make,talmor2019olmpics,schick2019ota,ettinger2020bert,kassner2019negated,sakaguchi2019winogrande}. \citet{schick2020exploiting} propose \pet{}, which combines this idea with gradient-based learning for efficient few-shot text classification. \citet{jiang2019know} and \citet{schick-etal-2020-automatically} consider the problem of finding the
best way to rephrase a given task as a cloze
question. \citet{schick2020just}'s version of \pet{} can generate  multiple tokens, but still requires a text classification objective and does not scale to long output sequences. \citet{radford2018language} consider task descriptions for text generation tasks, but do so only in a zero-shot setting. In a similar spirit, \citet{brown2020language} investigate the ability of pretrained language models to leverage task descriptions and examples without any gradient-based optimization.

\begin{figure*}
	\tikzset{
		every node/.style={
			outer sep=0, text height=1.5ex, text depth=0.25ex
		},
		input/.style={
			draw=c0, rounded corners, line width=1pt
		},
		pattern/.style={
			draw=c1, rounded corners, line width=1pt
		},
		label/.style={
			font=\sffamily\footnotesize, rounded corners, inner ysep=0.12cm, inner xsep=0.2cm, outer xsep=0.15cm, text=decentgrey!80!black, line width=1pt
		},
		arrow/.style={
			draw=decentgrey!80!black,->,>=latex
		},
	}
	\centering
	\begin{tikzpicture}
	\path[input] node[partialbox, font=\sffamily\footnotesize, fill=c0!10, outer sep=0, inner sep=0.15cm, align=center](input) {\textsf{American Duo Wins Opening Beach Volleyball Match}};
	
	\node[below=0.025cm of input.south, anchor=center, outer sep=0cm, inner sep=0cm, text=c0](input-label){\small $\mathbf{x}$};
	
	\node[font=\sffamily\footnotesize, left=0.15cm of input, inner sep=0, outer sep=0](pattern-text-1){News:};
	\node[font=\sffamily\footnotesize, left=0.15cm of pattern-text-1, inner sep=0, outer ysep=0.1cm](pattern-text-2){\mask{}};
	
	\begin{pgfonlayer}{bg}
	\path[pattern] node[partialbox=13pt, fit=(input)(pattern-text-1)(pattern-text-2), fill=c1!10, inner ysep=0.3cm, inner xsep=0.2cm](pattern){};
	\node[below=0.025cm of pattern.south, anchor=center, outer sep=0cm, inner sep=0cm, text=c1](pattern-label){\small $P(\mathbf{x})$};
	\end{pgfonlayer}
	
	\node[label, right=0.4cm of pattern](label-2){2};
	\node[label, above=0cm of label-2](label-1){1};
	\node[label, below=0cm of label-2, text=black, fill=c0!10, draw=c0, line width=0.5pt](label-3){3};
	\node[below=0.15cm of label-3, text=c0, inner sep=0](y-label){\small $y$};
	
	\node[label, right=0.4cm of label-2](verbalizer-2){World};
	\node[label, above=0cm of verbalizer-2.north west, anchor=south west](verbalizer-1){Business};
	\node[label, below=0cm of verbalizer-2.south west, anchor=north west, text=black, fill=c1!10, draw=c1, line width=0.5pt](verbalizer-3){Sports};
	\node[below=0.15cm of verbalizer-3, text=c1, inner sep=0](y-label){\small $v(y)$};
	
	\path[] (label-1) edge[arrow] (verbalizer-1);
	\path[] (label-2) edge[arrow] (verbalizer-2);
	\path[] (label-3) edge[arrow, draw=black] (verbalizer-3);
	
	\draw [black!75, dotted, thick, rounded corners, ->, >=latex] (verbalizer-3.east)--([xshift=0.4cm]verbalizer-3.east)--([xshift=0.4cm, yshift=1.7cm]verbalizer-3.east) -- ([yshift=1.7cm]verbalizer-3.east -| pattern-text-2.center) node [midway, fill=white] {\small $p(y \mid \mathbf{x}) \propto p_M(v(y) \mid P(\textbf{x}))$} -- (pattern-text-2.north);
	
	\end{tikzpicture}
	\caption{Application of a pattern-verbalizer pair $(P,v)$ in \pet{}: The input $\mathbf{x}$ is converted into a cloze question $P(\mathbf{x})$. The probability $p(y \mid \mathbf{x})$ of each label $y$ is derived from the probability that a pretrained model $M$ assigns to its verbalization $v(y)$ at the masked position. Figure adapted from \citet{schick-etal-2020-automatically}.}
	\label{figure:pet}
\end{figure*}
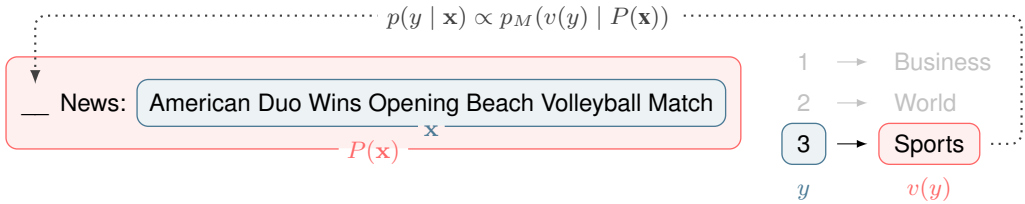

Other approaches to few-shot learning in NLP commonly require large sets of examples from related tasks \citep{gu-etal-2018-meta,dou-etal-2019-investigating,qian-yu-2019-domain,ye-etal-2020-zero}, parallel data for consistency training \citep{xie2019unsupervised,chen2020mixtext}, or highly specialized methods tailored towards a specific task \citep{laban-etal-2020-summary}. In contrast, \genpet{} requires no additional labeled data and provides an intuitive interface to leveraging task-specific human knowledge. 

Our work is also related to
prefix-constrained
decoding in interactive machine translation for making suggestions on how to complete a partial translation \citep{knowles2016neural,wuebker-etal-2016-models}.
\citet{keskar2019ctrl} and \citet{he2020ctrlsum} similarly use prompts
and keywords for controllable text generation, but require specific pretraining procedures and do so
only in high-resource settings.

\section{\pegasus{} Pretraining}
\label{sec:pegasus-pretraining}

We briefly summarize the pretraining procedure of \pegasus{} \citep{zhang2019pegasus}, the model to which we apply \genpet{}.
\pegasus{} is a standard Transformer encoder-decoder
architecture \citep{Vaswani2017} that is pretrained
using \emph{gap-sentence generation}, an objective tailored
to text summarization tasks. This pretraining objective
requires a set of documents consisting of multiple
sentences. The key idea is to preprocess each document by
(i) picking a subset of $m$ informative sentences,\footnote{The most informative sentences are selected
where informativeness is measured as the Rouge1 F1 score \citep{lin2004rouge} between the sentence and the remaining document.}
(ii) replacing each of these sentences by a mask token, and (iii) concatenating all removed sentences into a pseudo-summary. The Transformer model is then trained to generate this pseudo-summary given the partially masked document. 
Similar to prior
work \citep[e.g.,][]{raffel2019exploring,lewis2019bart},
this is done by
having the encoder 
process the entire masked document and the decoder generate the output autoregressively.

\citet{zhang2019pegasus} train two variants
of \pegasus{}: \pegasus{}-base, a 12-layer model with
approximately 223M parameters, and \pegasus{}-large, a
16-layer model with 568M parameters. As only the latter
version is publicly available in a variant that is not
finetuned on any downstream task, all our experiments are based on \pegasus{}-large.

\section{Pattern-Exploiting Training}
\label{sec:pet}

Pattern-Exploiting Training
(\pet{}, \citet{schick2020exploiting}) is a finetuning method for text classification tasks. That is, \pet{} can be applied to problems where a text sequence $\mathbf{x} \in \mathcal{X}$ must be mapped to a label $y$ from a finite set $\mathcal{Y}$. As shown in Figure~\ref{figure:pet}, \pet{} enables data-efficient text classification by converting inputs into cloze questions; this drastically reduces the number of examples required \citep{schick2020exploiting,schick2020just}.

Let $M$ be a masked
language model,
$V$ its vocabulary of tokens and  $\mask{} \in V$ the mask token; we denote the set of all token sequences as $V^*$. 
Given an input sequence $\mathbf{z} \in V^*$ that contains exactly one mask token, let $p_M(t \mid \mathbf{z})$ denote the probability assigned to $t \in V$ by $M$ at the masked position in $\mathbf{z}$. As illustrated in Figure~\ref{figure:pet}, \pet{} requires:
\begin{itemize}	\item a \emph{pattern} $P: \mathcal{X} \rightarrow V^*$ that maps each input 
$\mathbf{x}$
to a cloze question containing exactly one mask token;	\item a \emph{verbalizer} $v: \mathcal{Y} \rightarrow V$ that maps each 
label $y$
to a single token representing its meaning in the pattern. \end{itemize}The probability of $y$ given $\mathbf{x}$ is then derived from the probability that $M$ assigns to $v(y)$ at the masked position in $P(\mathbf{x})$:
\begin{equation}
p(y \mid \mathbf{x}) = \frac{p_M(v(y) \mid P(\mathbf{x}))}{ \sum_{y' \in \mathcal{Y}} p_M(v(y') \mid P(\mathbf{x}))}
\end{equation}
For finetuning, the cross-entropy between $p(y \mid \mathbf{x})$ and
the true label of $\mathbf{x}$ is used as training objective.

\section{Generation with Instructions}
\label{sec:gin}

We now introduce \genpet{}, our method for finetuning  language models with instructions for text \emph{generation}. Similar to \pet{}, we provide instructions by means of patterns  $P: \mathcal{X} \rightarrow V^*$ that we use to modify the original input. However, we do
not require a verbalizer as our output space already
consists of natural language sentences, i.e.,
$\mathcal{Y} \subseteq V^*$. 
In designing \genpet{}, we tackle
three key challenges
for few-shot text generation with instructions:
\begin{enumerate}[topsep=0.5em]
	\setlength\itemsep{-0.1em}
	\item How should we provide an instruction to an encoder-decoder model so that the model can make the best possible use of it? (\S\ref{sec:applying-a-single-pattern})
	\item How can we ensure that the model understands the instructions provided sufficiently well, and how do we deal with the fact that even minor modifications to the 
patterns can have a big impact on performance \citep{jiang2019know,schick2020exploiting,elazar2021measuring}? (\S\ref{sec:combining-patterns})	
\item How do we
prevent overfitting, a major issue in few-shot settings? (\S\ref{sec:preventing-overfitting})\end{enumerate}
\paragraph{Notation}
Let $P$ be a
pattern, $\mathbf{x} \in \mathcal{X}$ and
$\mathbf{y} \in \mathcal{Y}$
input and output text sequences,
and $\mathbf{z} = P(\mathbf{x})$
the result of applying $P$ to $\mathbf{x}$, i.e., a text sequence containing a single mask token. Furthermore, let  $\mathbf{y} = y_1 \ldots y_n$, $\mathbf{z} = z_1 \ldots z_m$ and let the mask token in $\mathbf{z}$ be at some position $h \leq m$. We denote the subsequence $y_i \ldots y_j$ by $\mathbf{y}_{i:j}$.

We consider an encoder-decoder model $M$ pretrained
by
masked language modeling.
That is, the model must
be able to compute a probability
$p_M(\mathbf{y} \mid \mathbf{z})$ that measures to what
extent $\mathbf{y}$ is a plausible substitute for the mask
in $\mathbf{z}$. We further require that this is done by
decomposing the joint probability of $\mathbf{y}$
as follows:\footnote{There are several recent architectures that meet this requirement, including \textsc{Bart} \citep{lewis2019bart}, T5 \citep{raffel2019exploring} and \pegasus{} \citep{zhang2019pegasus}.}
\begin{equation}
p_M(\mathbf{y} \mid \mathbf{z}) = \prod_{i=1}^n p_M(y_i \mid \mathbf{z}; \mathbf{y}_{1:i-1})
\end{equation}
where $p_M(y_i \mid \mathbf{z}; \mathbf{y}_{1:i-1})$ is
obtained
by processing $\mathbf{z}$ using the encoder and $\mathbf{y}_{1:i-1}$ using the decoder.
If we happen to already know some prefix $\mathbf{y}_{1:k-1}$ of $\mathbf{y}$, we denote with 
\begin{equation}
p_M(\mathbf{y}_{k:n} \mid \mathbf{z}; \mathbf{y}_{1:k-1}) = \prod_{i=k}^{n} p_M(y_i \mid \mathbf{z}; \mathbf{y}_{1:i-1})
\end{equation}
the probability that $M$ assigns to the remaining sequence $\mathbf{y}_{k:n}$ if the prefix $\mathbf{y}_{1:k-1}$ was already processed with the decoder.

\subsection{Using a Single Instruction}\label{sec:applying-a-single-pattern}

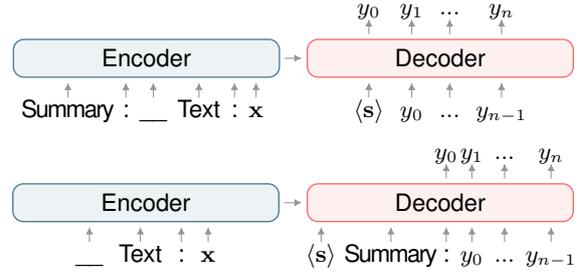
\begin{figure}
	\centering
	{\sffamily\footnotesize
		\tikzset{
			every node/.style={
				outer sep=0, text height=1.5ex, text depth=0.25ex,
			},
			input/.style={
				rounded corners, line width=1pt, font=\footnotesize, outer ysep=0, inner ysep=0.5pt,
			},
			model/.style={
				draw=c0, fill=c0!10, rounded corners, line width=0.5pt, minimum width=3.5cm, minimum height=0.5cm, outer sep=2pt,
			},
			fitter/.style={
				inner sep=0, outer sep=0,
			},
			arrow/.style={
				->,>={Latex[length=1mm, width=1mm]}, draw=darkgrey
			},
		}
		\begin{tikzpicture}
		
		\node[input, inner xsep=0](sum0){S\hspace*{-0.6pt}u\hspace*{-0.6pt}m\hspace*{-0.6pt}m\hspace*{-0.6pt}a\hspace*{-0.6pt}r\hspace*{-0.6pt}y};
		\node[input, right=0cm of sum0](col0){:};
		\node[input, inner xsep=1pt, right=0cm of col0](mas0){\mask{}};
		\node[input, right=0cm of mas0](tex0){Text};
		\node[input, inner xsep=1pt, right=0cm of tex0](c20){:};
		\node[input, right=0cm of c20](x0){\small$\mathbf{x}$};
		
		\node[fitter, fit=(sum0)(x0)](fitenc0){};
		\node[model, above=0.2cm of fitenc0](enc0){Encoder};	
		
		\node[input, right=0.8cm of x0 -| enc0.east](s0){$\langle\mathbf{s}\rangle$};
		\node[input, right=0cm of s0](t0-0){$y_0$};
		\node[input, right=0cm of t0-0](dots0){$...$};
		\node[input, right=0cm of dots0](tk0){$y_{n-1}$};
		
		\node[fitter, fit=(s0)(tk0)](fitdec0){};
		\node[model, draw=c1, fill=c1!10, above=0.2cm of fitdec0](dec0){Decoder};
		
		\node[input, above=0.2cm of dec0.north -| s0, anchor=south](t0out-0){$y_0$};
		\node[input, above=0.2cm of dec0.north -| t0-0, anchor=south](t1out-0){$y_1$};
		\node[input, above=0.2cm of dec0.north -| dots0, anchor=south](dotsout-0){$...$};
		\node[input, above=0.2cm of dec0.north -| tk0, anchor=south](tkout-0){$y_n$};
		
		\path[] (sum0) edge[arrow] (sum0 |- enc0.south);
		\path[] (col0) edge[arrow] (col0 |- enc0.south);
		\path[] (mas0) edge[arrow] (mas0 |- enc0.south);
		\path[] (tex0) edge[arrow] (tex0 |- enc0.south);
		\path[] (c20) edge[arrow] (c20 |- enc0.south);
		\path[] (x0) edge[arrow] (x0 |- enc0.south);
		
		\path[] (s0) edge[arrow] (s0 |- dec0.south);
		\path[] (t0-0) edge[arrow] (t0-0 |- dec0.south);
		\path[] (dots0) edge[arrow] (dots0 |- dec0.south);
		\path[] (tk0) edge[arrow] (tk0 |- dec0.south);
		
		\path[] (t0out-0 |- dec0.north) edge[arrow] (t0out-0);
		\path[] (t1out-0 |- dec0.north) edge[arrow] (t1out-0);
		\path[] (dotsout-0 |- dec0.north) edge[arrow] (dotsout-0);
		\path[] (tkout-0 |- dec0.north) edge[arrow] (tkout-0);
		
		\path[] (enc0) edge[arrow] (dec0);
		
		\node[fitter, below=-0.1cm of sum0](padding-bottom){};

		\end{tikzpicture}
		\begin{tikzpicture}
		
		\node[input](mas0){\mask{}};
		\node[input, right=0cm of mas0](tex0){Text};
		\node[input, right=0cm of tex0](c20){:};
		\node[input, right=0cm of c20](x0){\small$\mathbf{x}$};
		
		\node[fitter, fit=(mas0)(x0)](fitenc0){};
		\node[model, above=0.2cm of fitenc0](enc0){Encoder};
		
		\node[input, right=0.18cm of x0 -| enc0.east](s0){$\langle\mathbf{s}\rangle$};
		\node[input, right=0cm of s0, outer xsep=-3pt](sum0){S\hspace*{-0.6pt}u\hspace*{-0.6pt}m\hspace*{-0.6pt}m\hspace*{-0.6pt}a\hspace*{-0.6pt}r\hspace*{-0.6pt}y};
		\node[input, right=0cm of sum0, outer xsep=-1pt](col0){:};
		\node[input, right=0cm of col0, outer xsep=-1pt](t0-0){$y_0$};
		\node[input, right=0cm of t0-0, outer xsep=-1pt](dots0){$...$};
		\node[input, right=0cm of dots0, outer xsep=-1pt](tk0){$y_{n-1}$};
		
		\node[fitter, fit=(sum0)(tk0)](fitdec0){};
		\node[model, draw=c1, fill=c1!10, above=0.2cm of fitdec0, xshift=-0.3cm](dec0){Decoder};
		
		\node[input, above=0.2cm of dec0.north -| col0, anchor=south](t0out-0){$y_0$};
		\node[input, above=0.2cm of dec0.north -| t0-0, anchor=south](t1out-0){$y_1$};
		\node[input, above=0.2cm of dec0.north -| dots0, anchor=south](dotsout-0){$...$};
		\node[input, above=0.2cm of dec0.north -| tk0, anchor=south](tkout-0){$y_n$};
		
		\path[] (sum0) edge[arrow] (sum0 |- enc0.south);
		\path[] (col0) edge[arrow] (col0 |- enc0.south);
		\path[] (mas0) edge[arrow] (mas0 |- enc0.south);
		\path[] (tex0) edge[arrow] (tex0 |- enc0.south);
		\path[] (c20) edge[arrow] (c20 |- enc0.south);
		\path[] (x0) edge[arrow] (x0 |- enc0.south);
		
		\path[] (s0) edge[arrow] (s0 |- dec0.south);
		\path[] (t0-0) edge[arrow] (t0-0 |- dec0.south);
		\path[] (dots0) edge[arrow] (dots0 |- dec0.south);
		\path[] (tk0) edge[arrow] (tk0 |- dec0.south);
		
		\path[] (t0out-0 |- dec0.north) edge[arrow] (t0out-0);
		\path[] (t1out-0 |- dec0.north) edge[arrow] (t1out-0);
		\path[] (dotsout-0 |- dec0.north) edge[arrow] (dotsout-0);
		\path[] (tkout-0 |- dec0.north) edge[arrow] (tkout-0);
		
		\path[] (enc0) edge[arrow] (dec0);
		
		\node[fitter, left=1pt of enc0](padding-left){};
		
		\end{tikzpicture}
	}
	\caption{Generation process of an output $\mathbf{y}
		= y_0...y_n$ for input $\mathbf{x}$ when the
		instruction is entirely processed using the
		encoder (top) and when parts of it are processed
		using the decoder (bottom). We use
		$\langle\mathbf{s}\rangle$ to denote the model's
		start-of-sequence token.
		The seemingly subtle difference between the two setups can
		lead to quite different generations: 
		Instructions processed by the decoder
		have a stronger impact on the model’s predictions than those processed by the encoder.
	}
	\label{figure:decoder-prefix}
\end{figure}

As $M$ is an encoder-decoder language model,  
we have several options for how to apply a pattern $P$, i.e.,  how to ingest an instruction when computing the probability of $\mathbf{y}$ given $\mathbf{x}$: We may process the entire sequence $P(\mathbf{x}) = \mathbf{z}$ with the encoder, but we may also choose some index ${j < h}$ and process $\mathbf{z}_{1:j-1}\mathbf{z}_{h:n}$ using the encoder and $\mathbf{z}_{j:h-1}$ using the decoder. 
For example, if $\mathbf{z} =
\inlinepattern{\text{\footnotesize\sffamily{Summary: \mask{}
      Text: }}$\mathbf{x}$}$, we can process the prefix
``Summary:'' using the encoder or the decoder; that is, we may
compute either of the following (cf.\ Figure~\ref{figure:decoder-prefix}):
\begin{align}
p_1 & = p_M(\mathbf{y} \mid \pattern{\text{\footnotesize\sffamily{Summary: \mask{} Text: }}$\mathbf{x}$}) \label{prefix1}\\
p_2 & = p_M(\mathbf{y} \mid \pattern{{\footnotesize\sffamily{\mask{} Text: }}$\mathbf{x}$} ; \pattern{\footnotesize\sffamily{Summary:}})\label{prefix2}
\end{align}
In preliminary experiments, we found tokens that belong to
the partially generated output sequence (i.e., tokens that
are processed using the decoder) to have a much stronger impact
on the model's predictions than regular input tokens (i.e.,
those processed by the encoder). This applies all the more
to \pegasus{}, which is pretrained to always generate full
sentences:
If the pattern used consists of a partial sentence
(e.g., a short prompt) which is to be completed by the model, \pegasus{} tends to instead simply start a new sentence that does not relate to the given prefix if the latter is processed with the encoder.

Based on this observation, we supplement each pattern $P$ with a \emph{decoder prefix} $\mathbf{d} \in V^*$ that is given to the model as part of the generated sequence rather than the observed input. Accordingly, we define the probability of $\mathbf{y}$ given $\mathbf{x}$ as 
\begin{equation}
p_{(P, \mathbf{d})}(\mathbf{y} \mid \mathbf{x}) = p_M(\mathbf{y} \mid P(\mathbf{x}); \text{d})
\end{equation}
In Eqs.~\ref{prefix1} and~\ref{prefix2},
probability $p_1$ corresponds to using pattern
$P_1(\mathbf{x})
= \inlinepattern{\text{\footnotesize\sffamily{Summary: \mask{}
Text: }}$\mathbf{x}$}$ with an empty decoder prefix
$\mathbf{d}_1$, whereas $p_2$ corresponds to
using
the pattern
$P_2(\mathbf{x}) = \inlinepattern{{\footnotesize\sffamily{\mask{} Text: }}$\mathbf{x}$}$ with a decoder prefix $\mathbf{d}_2 = \inlinepattern{\footnotesize\sffamily{Summary:}}$. Both variants are illustrated in Figure~\ref{figure:decoder-prefix}.

We finetune $M$ on a set of training examples $(\mathbf{x}, \mathbf{y})$ simply by minimizing the cross-entropy between $p_{(P, \mathbf{d})}(\mathbf{y} \mid \mathbf{x})$ and $\mathbf{y}$ using teacher forcing.

\subsection{Combining Instructions}\label{sec:combining-patterns}

As shown in previous work \citep{jiang2019know,schick2020exploiting}, using different instructions or formulating the same input in different ways can have a strong impact on the model's performance. 
Unfortunately, in the absence
of a large development set, instructions that work well are often
hard to distinguish from those that perform poorly. We alleviate this issue by enabling the simultaneous usage of multiple instructions (represented by
multiple pairs of patterns and decoder prefixes) and combining them using a
mechanism similar to knowledge
distillation \citep{hinton2015distilling}. This mechanism
mitigates the negative influence of instructions that are
hard to understand for the model. This means that users can
simply provide \emph{all} (variants of) instructions that
they can think of. Further, it is much faster and more
memory efficient than having to constantly use multiple
instructions (and thus, multiple models) during
inference. \pet{} \citep{schick2020exploiting} also uses a
multi-pattern approach -- which is based on averaging the predictions
obtained with different patterns --, but it is not applicable in text generation settings as we cannot compute the average of multiple generated sequences in a meaningful way.

Given pairs of patterns and corresponding decoder prefixes $(P_1, \mathbf{d}_1), \ldots, (P_k, \mathbf{d}_k)$ and a set of models $M_1, \ldots, M_k$, where each $M_i$ was finetuned using $(P_i, \mathbf{d}_i)$, we aim to obtain a single model $\tilde{M}$ that contains the combined knowledge of all models. To do so, we require a small set of unlabeled examples $\mathcal{U}$. For each $\mathbf{x} \in \mathcal{U}$, we first generate one output sequence $\mathbf{y}^{(P_i, \mathbf{d}_i)}$ per $(P_i, \mathbf{d}_i)$ using greedy decoding as in \citet{zhang2019pegasus}, resulting in a set of candidate outputs 
$
\mathcal{C}_\mathbf{x} = \{ \mathbf{y}^{(P_i, \mathbf{d}_i)} \mid 1 \leq i \leq k \}
$.
To assign a score to each candidate $\mathbf{y} \in \mathcal{C}_\mathbf{x}$,
we first compute the log-likelihood of $\mathbf{y}$ for each $(P_i, \mathbf{d}_i)$ as
\begin{equation}
s_i(\mathbf{y} \mid \mathbf{x}) = \log p_{(P_i, \mathbf{d}_i)}(\mathbf{y} \mid \mathbf{x}) \label{s_i}
\end{equation}
The total score of $\mathbf{y}$ is then simply the exponentiated
average
over the patterns:
\begin{equation}
s(\mathbf{y} \mid \mathbf{x}) = \exp {\frac{1}{k} \sum_{i=1}^k s_i(\mathbf{y} \mid \mathbf{x})} \label{s_total}
\end{equation}
The model $\tilde{M}$ is trained on pairs $(\mathbf{x}, \mathbf{y})$ where $\mathbf{x} \in \mathcal{U}$ and $\mathbf{y}$ is drawn from $\mathcal{C}_\mathbf{x}$ with probability proportional to $s(\mathbf{y} \mid \mathbf{x})$.

While we could train this final model to simply maximize
 $p_{\tilde{M}}(\mathbf{y} \mid \mathbf{x})$, we note that
 this creates a large discrepancy between pretraining and
 finetuning: During pretraining, masked language models only process
 sequences that contain at least one mask token.
In the spirit of our intention to make pretraining and
 finetuning as similar as possible (\S\ref{intro}), we therefore train $\tilde{M}$ using a trivial pattern $P(\mathbf{x}) = \inlinepattern{\small\sffamily{\mask{}} $\mathbf{x}$}$ that just prepends a single mask token to the input and use an empty decoder prefix; that is, we maximize
$
p_{\tilde{M}}(\mathbf{y} \mid \inlinepattern{\small\sffamily{\mask{}} $\mathbf{x}$}; \inlinepattern{\ })
$
instead of $p_{\tilde{M}}(\mathbf{y} \mid \mathbf{x})$.
In addition to reducing the pretraining-finetuning
 discrepancy, putting the mask token \emph{before} the input
 biases the model towards generating text that is likely to
 precede the input. This is desirable because news
 articles -- which abound in big language models' pretraining data -- often have a headline and a short summary \emph{before} the article rather than after it.

\subsection{Preventing Overfitting}\label{sec:preventing-overfitting}

In preliminary experiments, we found pretrained encoder-decoder models to strongly
overfit the training data when trained on just a handful of
examples: When generating new texts, they often simply
reproduce phrases from training
examples, even if they are not in any way related to the
current input. To alleviate this issue, we introduce two
modifications to our training procedure; we
refer to them as \emph{unsupervised scoring} and \emph{joint
  training}.

\paragraph{Unsupervised Scoring} For unsupervised scoring, we compute $s(\mathbf{y} \,{\mid}\, \mathbf{x})$ as in Eq.~\ref{s_total}, but we use an \emph{untrained} model (i.e., one that has not been finetuned on task-specific examples) to compute $p_{(P_i, \mathbf{d}_i)} ( \mathbf{y} \,{\mid}\, \mathbf{x})$ in Eq.~\ref{s_i} for all $i \in \{1, \ldots, k\}$.

The intuition behind this is as follows: If for a given
input, a trained model simply reproduces phrases from its
training set, the resulting pair of
input and output texts should look strange to an untrained
model, which has not seen the example from which the output
is (partially) copied. Thus, sampling outputs from the
candidate set $\mathcal{C}_\textbf{x}$ based on the
probability assigned to each example by an untrained model
helps prevent overfitting: It results in the final model
being primarily trained on examples that also look natural
to a model that has not seen the training data.

We further use this idea to discard generated texts of really poor quality altogether. To this end, we sort the set $\mathcal{C} = \bigcup_{\mathbf{x} \in \mathcal{U}} \mathcal{C}_\textbf{x}$ of all outputs for all candidate sets based on their likelihood according to the untrained model in ascending order. Let the \emph{rank} $r_\textbf{y}$ of each output $\textbf{y} \in \mathcal{C}$ be its position in this sorted list, divided by the list's size. We then remove all outputs with $r_\textbf{y} < \tau$ from the candidate sets $\mathcal{C}_\textbf{x}$, where the threshold $\tau$ is a hyperparameter.

\paragraph{Joint Training} In \S\ref{sec:combining-patterns}, we assume the existence of an ensemble $\{M_1, \ldots, M_k\}$ where each model was trained using a different instruction. However, instead of training an individual
model $M_i$ for each pair $(P_i, \mathbf{d}_i)$, we can also
train a single model jointly on all instructions. To do so,
we simply replicate each training instance $k$ times and
process the $i$th copy with $(P_i, \mathbf{d}_i)$. Our
motivation is that forcing a single model to work well for
all instructions can act as a regularizer to prevent
overfitting. This approach comes with the additional
benefits of both being faster to train and generating less
overhead. Note that we still require
instruction combination 
(\S\ref{sec:combining-patterns})
because even given a single model understanding all
instructions, it would be unclear which instruction to
choose during test time, and querying the model with all
instructions would be inefficient.

\section{Experiments}
\label{sec:experiments}

\paragraph{Tasks} We evaluate \pegasus{} with and without
\genpet{} on a subset of the tasks in
\citet{zhang2019pegasus}. As our computing
resources are limited, we only choose those tasks for which
the
maximum
output length in \citet{zhang2019pegasus} is at most 128 tokens. We include the following tasks:
\begin{itemize}[topsep=0.5em]
	\setlength\itemsep{-0.1em}
\item \textbf{AESLC} \citep{zhang-tetreault-2019-aeslc}:
  Given an email body, predict the title of the email.
\item {\textbf{Gigaword} \citep{rush-etal-2015-neural}: Given
  the first sentence of a news article, generate its headline.}
\item \textbf{XSum} \citep{shashi-2018-xsum}: Summarize news articles spanning a wide range of different topics.
\item \textbf{Reddit TIFU} \citep{kim-etal-2019-reddittifu}:
  Generate summaries for posts from the TIFU community in Reddit.
\item \textbf{NEWSROOM} \citep{Grusky2018newsroom}: Generate
  summaries  for articles from various major publications.
\item \textbf{CNN/DailyMail} \citep{hermann2015cnndm}: For
  articles from CNN and the Daily Mail, generate a list of highlights.
\end{itemize}
For each task, we use the entire test set for
evaluation.\footnote{The only exception to this is NEWSROOM,
  which contains more than 100,000 examples:
  We only consider a
  subset of 10,000 examples
to ensure a
resource-friendly evaluation.} We create
two types of training sets containing either 10 or 100 training examples;
in addition, we provide 1,000 unlabeled examples per
task. Both unlabeled and training examples are obtained through uniform sampling
from each task's original training set.\footnote{\lsstyle We do not
reuse the datasets of \citet{zhang2019pegasus}
as they did not use a fixed seed and thus their
training data is not recoverable.}

As previous work \citep{schick2020just} has shown that the
choice of training examples has a large impact
on model performance, we create three distinct training
sets per size (10 and 100) and task using different random seeds, resulting in a total of six training sets per task. Scores
reported in this section are always average scores across
all three equal-sized sets of training examples, except for zero-shot settings where no training data is available at all.

\paragraph{Instructions} We use the same set of patterns across all tasks, but we combine them with different decoder prefixes. The patterns we use are:
\begin{align*}
P_1(\mathbf{x}) & = \pattern{\mask{} $\mathbf{x}$}  &
P_2(\mathbf{x}) & = \pattern{\mask{} {\footnotesize\sffamily{Text:}} $\mathbf{x}$} 
\end{align*}
All decoder prefixes are shown in
Table~\ref{table:generative-prefixes}. We combine each
pattern with each decoder prefix, resulting in four pairs
per task: $(P_1, d_1)$, $(P_1, d_2)$, $(P_2, d_1)$, $(P_2, d_2)$.

\begin{table}
	\footnotesize
	\setlength{\tabcolsep}{6pt}
	\begin{tabularx}{\linewidth}{lXX}
		\toprule
		\textbf{Task} & \multicolumn{2}{l}{\textbf{Decoder Prefixes}} \\
		\midrule
		AESLC &  $d_1\,{=}\,${\scriptsize\sffamily E-Mail Subject:} & $d_2\,{=}\,${\scriptsize\sffamily E-Mail Topic:} \\
		Gigaword &  $d_1\,{=}\,${\scriptsize\sffamily Headline:} & $d_2\,{=}\,${\scriptsize\sffamily Article Headline:} \\
		CNN/DM &  $d_1\,{=}\,${\scriptsize\sffamily Highlights:} & $d_2\,{=}\,${\scriptsize\sffamily Article Highlights:} \\
		Others &  $d_1\,{=}\,${\scriptsize\sffamily Short Summary:} & $d_2\,{=}\,${\scriptsize\sffamily Brief Summary:} \\
		\bottomrule
	\end{tabularx}
	\caption{Decoder prefixes we use for AESLC, Gigaword, CNN/DailyMail (CNN/DM) and all other summarization tasks (Others)}
	\label{table:generative-prefixes}
\end{table}

\begin{table*}
	\setlength{\tabcolsep}{2.8pt}
	\scriptsize
	\begin{tabularx}{\linewidth}{rXccccccc}
		\toprule
		\multicolumn{1}{c}{$t$} & Model & AESLC & Gigaword & XSum & Reddit TIFU & NEWSROOM & CNN/DailyMail & Avg \\
		\midrule
		\multirow{3}{*}{0} & \pegasus{} & {\p8.20}/{\p2.74}/{\p7.35} & {23.91}/{\p7.66}/{20.64} & {18.61}/{\p2.54}/{12.06} & {\bt 17.19}/{\bt \p3.29}/{\bt 12.00} & {23.24}/{11.20}/{18.34} & {\bt 35.20}/{\bt 14.07}/{22.84} & 21.06/\p6.91/15.54 \\
		& \pegasus{}-\textsc{m} & {12.39}/{\p4.74}/{11.42} & {19.63}/{\p5.51}/{16.97} & {\bt 32.43}/{\bt 13.10}/{\bt 24.58} & {14.80}/{\p2.89}/{10.74} & {25.01}/{13.57}/{20.90} & {33.36}/{12.97}/{22.63} & 22.94/\p8.80/17.87 \\
		& \genpet{} & {\bt 19.81}/{\bt \p8.81}/{\bt 18.53} & {\bt 28.01}/{\bt 10.48}/{\bt 24.92} & {29.24}/{10.56}/{22.73} & {15.41}/{\p2.83}/{11.63} & {\bt 26.35}/{\bt 15.79}/{\bt 23.22} & {33.08}/{12.82}/{\bt 23.27} & {\bt 25.32}/{\bt 10.21}/{\bt 20.71} \\
		\midrule
		\multirow{3}{*}{10} & \pegasus{} & {\p9.37}/{\p3.77}/{\p8.97} & {25.18}/{\p9.24}/{22.80} & {30.41}/{\p9.57}/{23.26} & {18.48}/{\p3.97}/{14.08} & {25.59}/{12.28}/{21.18} & {37.54}/{15.84}/{25.18} & 24.43/\p9.11/19.24 \\
		& \pegasus{}-\textsc{m} & {16.53}/{\p7.47}/{16.15} & {27.33}/{10.60}/{24.98} & {33.96}/{11.90}/{26.29} & {19.78}/{\p4.50}/{15.16} & {29.91}/{16.73}/{25.70} & {37.88}/{16.19}/{25.82} & 27.56/11.23/22.35 \\
		& \genpet{} & {\bt 27.19}/{\bt 14.08}/{\bt 26.73} & {\bt 30.93}/{\bt 13.02}/{\bt 28.49} & {\bt 35.88}/{\bt 13.22}/{\bt 28.24} & {\bt 22.43}/{\bt \p5.55}/{\bt 17.27} & {\bt 34.48}/{\bt 22.00}/{\bt 30.60} & {\bt 38.91}/{\bt 16.97}/{\bt 26.65} & {\bt 31.63}/{\bt 14.14}/{\bt 26.33} \\
		\midrule
		\multirow{3}{*}{100} & \pegasus{} & {23.22}/{10.24}/{22.43} & {30.80}/{12.27}/{27.92} & {40.23}/{16.68}/{31.90} & {24.24}/{\p6.28}/{18.72} & {33.13}/{20.24}/{28.80} & {39.64}/{16.94}/{26.79} & 31.87/13.77/26.10 \\
		& \pegasus{}-\textsc{m} & {25.87}/{12.34}/{24.99} & {31.38}/{12.65}/{28.33} & {40.73}/{17.10}/{32.43} & {24.74}/{\p6.40}/{19.10} & {34.79}/{21.60}/{30.37} & {\bt 40.08}/{17.14}/{27.06} & 32.93/14.54/27.05 \\
		& \genpet{} & {\bt 29.97}/{\bt 15.32}/{\bt 29.26} & {\bt 32.75}/{\bt 13.98}/{\bt 29.94} & {\bt 41.71}/{\bt 17.99}/{\bt 33.46} & {\bt 26.06}/{\bt \p7.34}/{\bt 20.34} & {\bt 36.20}/{\bt 23.51}/{\bt 32.02} & {40.02}/{\bt 17.77}/{\bt 27.79} & {\bt 34.45}/{\bt 15.98}/{\bt 28.80} \\
		\bottomrule
	\end{tabularx}
	
	\caption{R1/R2/RL scores for 
		six tasks and
		three training
		set sizes $t$; for 10 and 100 examples, all results
		are averaged across three different
		(seed-dependent)
		training sets. The last column shows average performance across all tasks.}
	\label{table:main-results}
\end{table*}

\paragraph{Setup}

For all our experiments with \genpet{}, we use \pegasus{}-large \citep{zhang2019pegasus} as underlying language model and perform greedy decoding; our implementation is based on the Transformers library \citep{wolf2019transformers} and PyTorch \citep{paszke2017automatic}.
Unless stated differently, all experiments are performed using the same setup as \citet{schick2020exploiting} and a single GPU with 11GB RAM (NVIDIA GeForce GTX 1080 Ti).

For optimizing hyperparameters, much previous few-shot work uses
development sets that are larger than the training sets by
multiple orders of
magnitude \citep[e.g.,][]{xie2019unsupervised,zhang2019pegasus,chen2020mixtext};
however, assuming the existence of such large development
sets is inconsistent with real-world few-shot
settings. In contrast, \citet{schick2020exploiting} assume
no development data at all and determine hyperparameters
based only on previous work and practical
considerations. We choose a middle course and create a small
development set of 100 examples for only one of the six
tasks, XSum. We use this development set in combination with
a single training set of 10 examples to determine
hyperparameters for \emph{all} tasks and training sets. However, we do so only for hyperparameters for which no consistent value can be derived from previous work.

Following \citet{zhang2019pegasus}, we use a maximum input
length of 512 tokens, the Adafactor
optimizer \citep{shazeer2018adafactor} with square root
learning rate decay, a dropout rate of 0.1 and label
smoothing setting
$\varepsilon=0.1$ \citep{szegedy2016rethinking}; we also
adopt \citet{zhang2019pegasus}'s
maximum output lengths for each task. As recommended by \citet{schick2020exploiting}, we train all models for 250 steps using a batch size of 8. We also tried training for 500 and 1,000 steps on our development set but found no major differences in performance. For the learning rate, we tried values of $\alpha\cdot10^{-5}$ with $\alpha \in \{ 1, 10, 50 \}$ as \citet{schick2020exploiting} use $\alpha = 1$ and \citet{zhang2019pegasus} use $\alpha = 50$; we found $\alpha = 10$ to perform best for all models. 
For unsupervised scoring (\S\ref{sec:preventing-overfitting}), we use a threshold of $\tau = 0.2$, i.e., we discard the 20\% of examples that are least likely according to an untrained model. We chose this value by looking at texts generated by \pegasus{} trained on 10 examples from the XSum development set, where we found the bottom 20\% to contain texts of poor quality, including random telephone numbers and repetitions of the same word.
For evaluation, we follow \citet{zhang2019pegasus} and report Rouge1, Rouge2 and RougeL (R1/R2/RL) F1 scores \citep{lin2004rouge} after stemming using the Porter algorithm \citep{porter1980algorithm}.

\paragraph{Results} On all six tasks, we compare the following three approaches for finetuning a pretrained \pegasus{} model:
\begin{itemize}
	\item \pegasus{}: The regular finetuning procedure described in \citep{zhang2019pegasus}.
	\item \pegasus{}-\textsc{m}: Finetuning with a
          single trivial pattern that inserts a mask token
          before the first word.
	\item \genpet{}: Finetuning with \genpet{} using
          patterns $P_1$ and $P_2$ and the decoder prefixes
          in Table \ref{table:generative-prefixes} as described above; we apply all modifications described in \S\ref{sec:preventing-overfitting}.
\end{itemize}
We do not compare to other
few-shot approaches as they either make quite different assumptions -- for example, \genpet{} requires manually designed patterns and some amount of unlabeled examples, whereas meta learning approaches \citep[e.g.,][]{gu-etal-2018-meta,dou-etal-2019-investigating,qian-yu-2019-domain} require large annotated datasets for related tasks --, or they cannot be transferred to a generative setting in a straightforward fashion, as is the case for consistency-based methods such as those of \citet{xie2019unsupervised} and \citet{chen2020mixtext}. However, we note that \pegasus{} is a strong baseline in terms of data efficiency, almost matching the performance of prior state-of-the-art systems trained on the full datasets with as little as 100 examples for many tasks \citep{zhang2019pegasus}.

Table~\ref{table:main-results} shows results for
zero-shot learning and for few-shot learning with 10 and 100
training examples. In the few-shot settings, \genpet{}
consistently outperforms \pegasus{} across all tasks,
resulting in an average improvement in R1 over \pegasus{} of
7.20 (31.63 vs 24.43) and 2.58 (34.45 vs 31.87).
\pegasus{}-\textsc{m} performs better than regular finetuning, indicating 
that even just adding a single mask token at the very beginning, 
without any instructions,
already  effectively  improves
performance.
(Recall that the effect of the initial mask is
to make finetuning more similar to pretraining
and to bias the models towards generating text
that is likely to appear before the input;
see \S\ref{sec:combining-patterns}).
However, it still performs clearly worse
than 
\genpet{}, demonstrating that
\pegasus{}
is indeed able to make use of the instructions
provided. 
In
the zero-shot setting, \genpet{} also outperforms all
baselines on average, but falls short on individual tasks.

\paragraph{Quantitative Analysis} To analyze  the factors contributing
to \genpet{}'s performance,
Table~\ref{table:ablation} 
compares the performance of the best (``best only'') and the
worst
(``worst only'')
performing pairs of pattern and decoder prefix to that of \genpet{} in a
setting with 10 training examples. We see some difference in
performance between using only the best and worst pairs,
but this difference is not as pronounced as in previous
work \citep{schick2020just,schick2020exploiting} -- possibly because our instructions are more similar to each other than patterns in prior work. Notably,
our strategy for combining instructions clearly performs
better than using just the best instruction across all tasks and
measures (compare \genpet{} with ``best only'').
Table~\ref{table:ablation} also shows results for 
using the best pattern without a decoder prefix (``no
dec.\ prefix'') and instead processing the entire input
using the encoder. That is, given $(P, \mathbf{d})$ with
$P(\mathbf{x}) = z_1\ldots z_n$ and $z_h = \mask{}$, we
compute $p_M(\mathbf{y} \mid z_1 \ldots z_{h-1} \mathbf{d}
z_h \ldots z_n)$ rather than $p_M(\mathbf{y} \mid z_1\ldots
z_n ; \mathbf{d})$ similar to the example shown in Figure~\ref{figure:decoder-prefix} (top).
While this variant still performs better than
\pegasus{}-\textsc{m} on two out of three datasets, results
clearly show that \pegasus{} makes less use of task
descriptions if they are processed using the encoder.

The bottom two rows of Table~\ref{table:ablation} show
performance when we replace unsupervised scoring
(\S\ref{sec:preventing-overfitting}) with regular scoring
using the supervised models (``sup. scoring'') and if we
additionally do not perform joint training (``no joint
train.''). As can be seen, not using joint training hurts
performance for all three tasks and supervised scoring hurts performance for two out of three tasks.

\begin{table}
	\setlength{\tabcolsep}{1.9pt}
	\scriptsize
	\begin{tabularx}{\linewidth}{Xccc}
		\toprule
		Model & AESLC & XSum & NEWSROOM \\
		\midrule
		\pegasus{} & {\p9.37}/{\p3.77}/{\p8.97} & {30.41}/{\p9.57}/{23.26} & {25.59}/{12.28}/{21.18} \\
		\pegasus{}-\textsc{m} & {16.53}/{\p7.47}/{16.15} & {33.96}/{11.90}/{26.29} & {29.91}/{16.73}/{25.70} \\
		\midrule
		\genpet{} & {\bt 27.19}/{\bt 14.08}/{\bt 26.73} & {\bt 35.88}/{\bt 13.22}/{\bt 28.24} & {\bt 34.48}/{22.00}/{\bt 30.60} \\
		\ └ worst only & {24.08}/{12.22}/{23.58} & {33.85}/{11.95}/{26.60} & {32.55}/{19.73}/{28.59} \\
		\ └ best only & {24.80}/{12.48}/{24.19} & {34.15}/{12.05}/{26.78} & {33.94}/{21.34}/{30.03} \\
		\quad └ no dec. prefix & {15.49}/{\p7.24}/{15.09} & {34.12}/{11.95}/{26.41} & {32.56}/{20.15}/{28.64} \\
		\ └ sup. scoring & {25.33}/{13.41}/{24.87} & {35.68}/{13.19}/{28.06} & {34.37}/{\bt 22.04}/{30.53} \\
		\quad └ no joint train. & {24.37}/{12.67}/{24.00} & {35.41}/{13.15}/{27.95} & {34.04}/{21.95}/{30.35} \\
		\bottomrule
	\end{tabularx}
	\caption{R1/R2/RL scores for several baselines and variants of \genpet{} given 10 training examples}
	\label{table:ablation}
\end{table}

\paragraph{Qualitative Analysis}

\begin{table}
	\footnotesize
	\setlength{\tabcolsep}{2pt}
	\begin{tabularx}{\linewidth}{lX}
		\toprule
		\multicolumn{2}{p{0.97\linewidth}}{\textbf{Input}: the dollar slipped against the euro on friday after the u.s. federal reserve cut its discount rate to banks by a half percentage point.} \\
		\midrule
		\textsc{Pg} & federal reserve cut its discount rate to banks by a half percentage point. \\
		\textsc{Pg}-\textsc{m} & \lsstyle The dollar fell against the euro on monday after the u.s. \\
		\genpet{} & dollar slips against euro after federal reserve cuts discount rate to banks. \\
		Gold & dollar slides against euro as fed cuts discount rate \\
		\bottomrule
	\end{tabularx}
	\caption{Zero-shot
          summaries
for the news item given as ``\textbf{Input}''.
\pegasus{} (\textsc{Pg}) simply creates a verbatim copy of the second part
of the input. \pegasus{}-\textsc{m} (\textsc{Pg}-\textsc{m}) hallucinates (``Monday''
vs.\ ``Friday''). \genpet{}'s summary is close in quality to
gold.}
	\label{table:qa-zeroshot}
\end{table}

Table~\ref{table:qa-zeroshot} shows
zero-shot abilities of three methods for one selected
input from Gigaword that illustrates some typical behaviors:
Regular \pegasus{}
just creates a verbatim copy of the input's second half --
this is true not only for this particular example, but can
be seen frequently for all datasets. We
assume this is due to the fact that \citet{zhang2019pegasus}
introduce some modifications to their training procedure
that encourage the model to copy
text. \pegasus{}-\textsc{m} is able to produce an output
that is not just a word-for-word copy of the input, but
hallucinates information that is not
backed by
the input text (``monday'').
We found that hallucination is a frequent problem for \pegasus{}-\textsc{m}.
This is hardly surprising given that the model has no
way of knowing that it is expected to generate a factual
headline summarizing the  input. In contrast, \genpet{} generates a fluent and factual headline that covers all relevant aspects.

\section{Conclusion}
\label{sec:conclusion}

We investigated the ability of pretrained language
models to make use of simple instructions with the aim of enabling more data-efficient text generation. 
We identified three major
challenges:  enabling language models to make good use of
the instructions provided, ensuring that the 
instructions  are useful and preventing overfitting. We
tackle these in our proposed approach, \genpet{}, by (i)
introducing the concept of decoder prefixes, (ii) combining
instructions
through knowledge distillation where target
sequences are generated with
probabilistically sampled
instructions
and
(iii) making use of unsupervised scoring and joint training. A pretrained \pegasus{} model finetuned with \genpet{} clearly outperforms regular finetuning in few-shot settings. 

\paragraph*{Acknowledgments}
This work was funded by the European Research Council (ERC \#740516).
We thank the anonymous reviewers
for their helpful comments.

\bibliography{literatur}
\bibliographystyle{acl_natbib}

\clearpage
\appendix

\section{Analysis}

\paragraph{Sequence Length}

We look at the performance of \genpet{} as a function
of the maximum output length $\ell$. One might be concerned that the
influence of the decoder prefix on generated tokens may
decrease with distance. This would mean that diminishing
gains are to be expected from \genpet{} for tasks that require
longer text sequences to be generated. To investigate
whether this is a problem for \genpet{}, Table~\ref{table:sequencelength} shows the
performance of \pegasus{} and \genpet{} for
all tasks with an original maximum output length of 128 tokens, using maximum output lengths of $\ell = 32$ and
$128$. 

For both values of $\ell$, we compute the gains $g_\ell$
from using \genpet{} as the difference in performance
between \genpet{} and \pegasus{}.  On average, increasing
$\ell$ to 128 tokens reduces the gains from \genpet{} over
regular finetuning by just $g_{32} - g_{128} = 0.10$ points
R1. This shows that instructions provided using \genpet{}
have a strong impact on generated tokens even if there are
dozens of other tokens in between. Thus,
\genpet{} works not only for short sequences, but
is also beneficial for generating long
text sequences.

\paragraph{Unsupervised Scoring}

We motivated the use of unsupervised scoring in
Section~5.2 by the observation that
\pegasus{} tends to
overfit the training data. This can for example be seen when
training \pegasus{} with individual instructions on the 10
examples from the XSum dataset used to optimize
hyperparameters. One of these examples has the gold-standard
summary ``Hugo Chavez [\ldots] is one of the most visible,
vocal and controversial leaders in Latin America''; as shown
in Table~\ref{table:qa-overfitting}, this induces
\pegasus{} to generate the phrase ``the most visible, vocal and controversial'' for many other inputs, even in cases where this phrase does not make any sense given the input text. Out of the summaries generated for 1,000 unlabeled examples, we found 92 to contain this particular phrase word-for-word.

Table~\ref{table:qa-overfitting} also shows the rank of each output as defined in Section~5.3 (i.e., its relative position in a list of all generated outputs that is sorted by likelihood in ascending order) both when likelihood is assigned using the trained models ($r_\text{sup}$) and when it is assigned using a fully unsupervised \pegasus{} model ($r_\text{unsup}$). As can be seen, an untrained model indeed assigns much less likelihood to those examples, thus downweighting their influence on the final model. For example, the last text shown in Table~\ref{table:qa-overfitting} is more probable than 92\% of all generated texts according to the trained model, compared to 24\% for the untrained model. With unsupervised scoring, the first three examples shown are even completely removed from the training set for the final model as their rank is below the chosen threshold of $\tau = 0.2$.

\begin{table}
	\setlength{\tabcolsep}{3pt}
	\scriptsize
	\begin{tabularx}{\linewidth}{cXccc}
		\toprule
		$\ell$ & Model & Reddit TIFU & NEWSROOM & CNN/DailyMail \\
		\midrule
		\multirow{2}{*}{$2^7$} & \pegasus{} & {18.48}/{\p3.97}/{14.08} & {25.59}/{12.28}/{21.18} & {37.54}/{15.84}/{25.18} \\
		& \genpet{} & {22.43}/{\bt \p5.55}/{17.27} & {\bt 34.48}/{\bt 22.00}/{\bt 30.60} & {\bt 38.91}/{\bt 16.97}/{\bt 26.65} \\
		\midrule
		\multirow{2}{*}{$2^5$} & \pegasus{} & {18.76}/{\p3.97}/{14.36} & {24.71}/{11.41}/{20.49} & {31.81}/{13.16}/{22.69} \\
		& \genpet{} & {\bt 22.45}/{\p5.54}/{\bt 17.32} & {33.89}/{21.26}/{30.02} & {33.44}/{14.35}/{24.17} \\
		\bottomrule		
	\end{tabularx}
	\caption{R1/R2/RL scores with maximum output lengths of $2^5=32$ and $2^7=128$ given 10 training examples}
	\label{table:sequencelength}
\end{table}

\begin{table*}
	\footnotesize
	\setlength{\tabcolsep}{3pt}
	\begin{tabularx}{\linewidth}{Xcc}
		\toprule
		Text & $r_\text{sup}$ & $r_\text{unsup}$ \\
				\midrule
		Margaret Thatcher, [\ldots] was \textbf{one of the most visible, vocal and controversial} leaders in the world. & 0.77 & 0.19 \\
		Bruce Forsyth [\ldots] was \textbf{one of the most visible, vocal and controversial} entertainers in the business. & 0.51 & 0.18 \\
		{[\ldots]} Hawaii Five-O, a police drama that was \textbf{one of the most visible, vocal and controversial} of all-time. & 0.41 & 0.11 \\
		Mongolia is \textbf{one of the most visible, vocal and controversial} countries in the world. & 0.81 & 0.32 \\
		The state pension is \textbf{one of the most visible, vocal and controversial} of all-time. & 0.92 & 0.24 \\
		\bottomrule
	\end{tabularx}
	\caption{Texts generated by \pegasus{} trained with
		individual patterns using \genpet{} on an XSum
		training set.
		Each of the five texts contains a phrase
		(highlighted in bold)
		from
		one specific
		training example.
		The right columns show the
		(normalized) rank of each output both with
		supervised scoring ($r_\text{sup}$) and
		unsupervised scoring ($r_\text{unsup}$).
		In these five examples, unsupervised scoring more
		effectively identifies the ``parroted''
		phrase as not being a good fit for  its new context.}
	\label{table:qa-overfitting}
\end{table*}

\paragraph{Variance}

To quantify the significance of performance improvements with \genpet{} over our two baselines, \pegasus{} and \pegasus{}-\textsc{m}, Table~\ref{table:variance} shows the standard deviation of Rouge1/Rouge2/RougeL scores across the three different training sets for all tasks considered.

\begin{table*}
	\scriptsize
	\begin{tabularx}{\linewidth}{rlYYY}
		\toprule
		$|T|$ & Model & AESLC & Gigaword & XSum \\
		\midrule
		\multirow{3}{*}{10} & \pegasus{} & {\p9.37}{$\pm$2.08} / {\p3.77}{$\pm$1.07} / {\p8.97}{$\pm$2.17} & {25.18}{$\pm$0.77} / {\p9.24}{$\pm$0.41} / {22.80}{$\pm$0.61} & {30.41}{$\pm$0.44} / {\p9.57}{$\pm$0.27} / {23.26}{$\pm$0.29} \\
		& \pegasus{}-\textsc{m} & {16.53}{$\pm$1.73} / {\p7.47}{$\pm$0.95} / {16.15}{$\pm$1.73} & {27.33}{$\pm$0.51} / {10.60}{$\pm$0.34} / {24.98}{$\pm$0.46} & {33.96}{$\pm$1.52} / {11.90}{$\pm$1.09} / {26.29}{$\pm$1.53} \\
		& \genpet{} & {\bt 27.19}{$\pm$1.93} / {\bt 14.08}{$\pm$1.13} / {\bt 26.73}{$\pm$1.99} & {\bt 30.93}{$\pm$0.15} / {\bt 13.02}{$\pm$0.17} / {\bt 28.49}{$\pm$0.17} & {\bt 35.88}{$\pm$1.42} / {\bt 13.22}{$\pm$1.17} / {\bt 28.24}{$\pm$1.50} \\
		\midrule
		\multirow{3}{*}{100} & \pegasus{} & {23.22}{$\pm$0.29} / {10.24}{$\pm$0.46} / {22.43}{$\pm$0.28} & {30.80}{$\pm$0.52} / {12.27}{$\pm$0.50} / {27.92}{$\pm$0.49} & {40.23}{$\pm$0.10} / {16.68}{$\pm$0.10} / {31.90}{$\pm$0.06} \\
		& \pegasus{}-\textsc{m} & {25.87}{$\pm$0.06} / {12.34}{$\pm$0.11} / {24.99}{$\pm$0.13} & {31.38}{$\pm$0.05} / {12.65}{$\pm$0.19} / {28.33}{$\pm$0.12} & {40.73}{$\pm$0.06} / {17.10}{$\pm$0.03} / {32.43}{$\pm$0.04} \\
		& \genpet{} & {\bt 29.97}{$\pm$0.39} / {\bt 15.32}{$\pm$0.36} / {\bt 29.26}{$\pm$0.54} & {\bt 32.75}{$\pm$0.26} / {\bt 13.98}{$\pm$0.09} / {\bt 29.94}{$\pm$0.16} & {\bt 41.71}{$\pm$0.06} / {\bt 17.99}{$\pm$0.02} / {\bt 33.46}{$\pm$0.08} \\
		\bottomrule
		&&&&\\
		\toprule
		$|T|$ & Model & Reddit TIFU & NEWSROOM & CNN/DailyMail \\
		\midrule
		\multirow{3}{*}{10} & \pegasus{} & {18.48}{$\pm$0.85} / {\p3.97}{$\pm$0.26} / {14.08}{$\pm$0.41} & {25.59}{$\pm$1.07} / {12.28}{$\pm$1.29} / {21.18}{$\pm$1.11} & {37.54}{$\pm$0.39} / {15.84}{$\pm$0.27} / {25.18}{$\pm$0.27} \\
		& \pegasus{}-\textsc{m} & {19.78}{$\pm$1.44} / {\p4.50}{$\pm$0.39} / {15.16}{$\pm$0.84} & {29.91}{$\pm$0.29} / {16.73}{$\pm$0.37} / {25.70}{$\pm$0.26} & {37.88}{$\pm$0.63} / {16.19}{$\pm$0.34} / {25.82}{$\pm$0.23} \\
		& \genpet{} & {\bt 22.43}{$\pm$0.78} / {\bt \p5.55}{$\pm$0.30} / {\bt 17.27}{$\pm$0.30} & {\bt 34.48}{$\pm$0.74} / {\bt 22.00}{$\pm$0.70} / {\bt 30.60}{$\pm$0.71} & {\bt 38.91}{$\pm$0.56} / {\bt 16.97}{$\pm$0.19} / {\bt 26.65}{$\pm$0.14} \\
		\midrule
		\multirow{3}{*}{100} & \pegasus{} & {24.24}{$\pm$0.32} / {\p6.28}{$\pm$0.01} / {18.72}{$\pm$0.27} & {33.13}{$\pm$0.47} / {20.24}{$\pm$0.80} / {28.80}{$\pm$0.48} & {39.64}{$\pm$0.13} / {16.94}{$\pm$0.16} / {26.79}{$\pm$0.18} \\
		& \pegasus{}-\textsc{m} & {24.74}{$\pm$0.08} / {\p6.40}{$\pm$0.05} / {19.10}{$\pm$0.01} & {34.79}{$\pm$0.55} / {21.60}{$\pm$0.74} / {30.37}{$\pm$0.54} & {\bt 40.08}{$\pm$0.23} / {17.14}{$\pm$0.09} / {27.06}{$\pm$0.07} \\
		& \genpet{} & {\bt 26.06}{$\pm$0.07} / {\bt \p7.34}{$\pm$0.09} / {\bt 20.34}{$\pm$0.12} & {\bt 36.20}{$\pm$0.56} / {\bt 23.51}{$\pm$0.69} / {\bt 32.02}{$\pm$0.54} & {40.02}{$\pm$0.22} / {\bt 17.77}{$\pm$0.07} / {\bt 27.79}{$\pm$0.05} \\
		\bottomrule
	\end{tabularx}
	\caption{Average R1/R2/RL scores and standard deviation ($\pm$) for 
		10 and 100 training examples across three different
		(seed-dependent)
		training sets.}
	\label{table:variance}
\end{table*}

\end{document}